\documentclass[11pt,preprint]{elsarticle}
\usepackage[utf8]{inputenc}
\usepackage{amsmath}
\usepackage{amsthm}
\usepackage{amssymb}
\usepackage[bookmarks=false,
 breaklinks=true,pdfborder={0 0 0},pdfborderstyle={},backref=section,colorlinks=false]
 {hyperref}

\makeatletter

\DeclareTextSymbolDefault{\textquotedbl}{T1}

\theoremstyle{plain}
\newtheorem{thm}{\protect\theoremname}
\theoremstyle{plain}
\newtheorem{lem}[thm]{\protect\lemmaname}

\usepackage{url}
\usepackage{xurl}

\makeatother

\providecommand{\lemmaname}{Lemma}
\providecommand{\theoremname}{Theorem}

\begin{document}
\begin{frontmatter}
\title{Why Shallow Neural Networks Learn Sinusoidal Components First: An
Orthogonal Polynomial Dynamics Framework}
\author[siat]{Yanyang Xiao\corref{cor1}}
\cortext[cor1]{Corresponding author}
\ead{xyy82148@gmail.com}
\address[siat]{Shenzhen Institutes of Advanced Technology, Chinese Academy of Sciences}
\begin{abstract}
Neural networks often learn low-frequency components before high-frequency
ones -{}-{}- a phenomenon called spectral bias or termed F-principle.
Many numerical studies have demonstrated this phenomenon, and theoretical
studies have sought to explain it. However, several goals remain insufficiently
addressed: (1) explaining spectral bias asymptotically, including
its origin and why the preferred components are generally sinusoidal;
(2) extending the explanation to general activation functions, such
as sigmoid and ReLU; (3) making weak assumptions about the data distribution;
(4) avoiding aliasing effects, such as those arising in finite-domain
Fourier analysis; and (5) developing a theoretical framework applicable
to arbitrary networks. We address these goals theoretically by analyzing
the eigenvalue decay and eigenvector shapes of the dynamics matrix
$H$, which arises when we expand the network output as a general
Fourier series (GFS) in an orthogonal polynomial basis and express
the gradient-descent dynamics of the GFS coefficients as ordinary
differential equations. The graded structure of $H$ enables an intuitive
explanation of spectral bias, while orthogonal polynomial theory helps
us analyze how learning bias depends on the activation function and,
potentially, the network architecture.
\end{abstract}
\begin{keyword}
artificial neural network \sep spectral bias \sep orthogonal polynomials
\sep singular value decomposition \sep asymptotic eigenvector
\end{keyword}
\end{frontmatter}

\section{Introduction}

In light of the Universal Approximation Theorem for neural networks,
there is a shadow over how networks can learn useful things. Along
this path, subjects such as learning bias, implicit regularization,
and information bottlenecks arise. In particular, a phenomenon called
spectral bias, or the F-principle, was discovered and was thought
to be connected to implicit regularization, to have explanatory power
for early stopping and generalization, and to expose weaknesses on
high-frequency tasks \citep{xu_frequency_2020,basri_frequency_2020,rahaman_onthespectral_2019}.
However, the observations in \citep{xu_frequency_2020}, although
clear in deriving the main conclusion, relied on finite-domain Fourier
analysis, in which lower-amplitude frequency components are largely
contaminated by aliasing effects. To fix this flaw and provide a quantitative
theory to explain it was the motivation of this work. Specifically,
we would like to (1) find an aliasing-free representation to describe
the frequency components; (2) provide the dynamics (time-dependent
evolution) of the frequency components in that representation; (3)
explain the spectral bias phenomenon through these dynamics; (4) allow
the explanation to be only asymptotic, leaving room for easier computation,
easier interpretation, and intuitive understanding; and (5) ensure
that the framework (constituted by 1--4) has the potential to extend
to networks beyond toy models.

Over the years, various theoretical works have emerged. The Neural
Tangent Kernel (NTK) theory was the first fundamental theory closely
connected to our work, in which, in principle, the training dynamics
of any network can be written as evolution through the kernel \citep{jacot_neuraltangent_2018},
yet the computation of the kernel involves complex integration. Some
recent papers accomplish impressive rigorous computations; for example,
\citep{murray_characterizing_2023} connect the decay speed of Hermite
coefficients to NTK coefficients and network depth, and \citep{zhang_whyshallow_2025}
obtain the NTK analytically through integration for ReLU, so that
various properties can be derived. However, \citep{murray_characterizing_2023}
do not discuss eigenvector shape, and \citep{zhang_whyshallow_2025}
discuss the shape of eigenvectors but only for the ReLU activation
function. In contrast, our theory applies to any sensible activation
function and, when explaining the shape of eigenvectors, requires
no closed-form integral.

A wide range of natural signals, such as natural images, natural sounds,
and EEG recordings, exhibit low-frequency domination. Traditional
ways of dealing with them often rely on or connect to Fourier analysis.
The sinusoidal shape of the \textquotedbl frequency components\textquotedbl{}
in the frequency bias happens to coincide with the \textquotedbl principal
components\textquotedbl{} of those natural signals. It may be because
of this coincidence that neural networks happen to be good at processing
natural signals. We would like to provide a mathematical perspective
on this as a by-product of this work.

The paper is organized as follows: First, we re-represent the output
of the network using an Orthogonal-Polynomial-based General Fourier
Series (GFS). Then, we derive the dynamics of the GFS coefficients,
which constitute the main governing equation. To analyze the dynamics
of the governing equation, we then explore the key properties of the
dynamics matrix $H$---namely, its eigenvalues and eigenvectors---using
a shallow neural network as an example. Finally, we discuss the relation
of the $H$ matrix to the loss landscape problem and effective dimension.

\section{General Fourier Series Expansion Derived from Loss}

Let's say we have target function $y(x)$, $\mathbb{R}\mapsto\mathbb{R}$.
The problem setup is to fit $J$ given pairs of feature and label
(inputs and outputs) $\left\{ \left(x_{j},y_{j}\right)\right\} $
by a function $\Upsilon_{\theta}(x)$ with parameter $\theta$ (usually
a vector). Without loss of generality, we limit the range of $x_{j}$
to $[-1,1]$. Generalize to high dimensional $x_{j}\in[-1,1]^{d}$
is under consideration.

We have the 2-norm loss for the error of the fitting:

\begin{equation}
\tilde{L}_{J}(\theta)=\frac{1}{2J}\sum_{j=1}^{J}\left(\Upsilon_{\theta}(x_{j})-y_{j}\right)^{2}.\label{eq:Loss-sample}
\end{equation}

Assume the samples $\left\{ x_{j}\right\} $ are draw from a distribution
$\rho(x)$, and the labels are come from an underlying function $y(x)$:
$y_{j}=y(x_{j})$. Now let $J\rightarrow\infty$, we should have a
theoretical loss $L(\theta)$:
\begin{equation}
L(\theta)=\frac{1}{2}\int_{\Omega}\left(\Upsilon_{\theta}(x)-y(x)\right)^{2}\rho(x)\,\mathrm{d}x.\label{eq:Loss}
\end{equation}
\[
\tilde{L}_{J}(\theta)\rightarrow L(\theta)\quad\mathrm{as}\ J\rightarrow\infty,
\]

The Eq. \eqref{eq:Loss} naturally induced a norm and an inner product
(assume all real functions)
\begin{equation}
\left\Vert f\right\Vert _{\rho}=\int_{\Omega}f^{2}(x)\rho(x)\,\mathrm{d}x,
\end{equation}
\begin{equation}
\left\langle f,g\right\rangle _{\rho}=\int_{\Omega}f(x)g(x)\rho(x)\,\mathrm{d}x.\label{eq:inner-product}
\end{equation}

Under the inner product in Eq. \eqref{eq:inner-product}, there exist
a set of orthonomal polynomial basis $\left\{ p_{k}(x)\right\} _{k=0}^{\infty}$
where $p_{k}(x)$ is a polynomial of order $k$:
\begin{equation}
\left\langle p_{k},p_{j}\right\rangle _{\rho}=\delta_{kj}.
\end{equation}

In 1-D case, each $p_{k}(x)$ has exactly $k$ zeros. Therefore the
subscript $k$ has similar meaning as the $k$ in Fourier mode $e^{\mathrm{i}kx}$,
i.e. the frequency.

If $\rho(x)>0$ almost everywhere, then $\left\{ p_{k}(x)\right\} _{k=0}^{\infty}$
is a complete basis with the same convergence trends as Chebyshev
approximation.

The fitting function $\Upsilon_{\theta}$, assumed piecewise continuous,
can be expanded in Generalized Fourier Series (GFS):
\begin{equation}
\Upsilon_{\theta}(x)=\sum_{k=0}^{\infty}c_{k}(\theta)p_{k}(x)\quad\textrm{(a.e.)}\label{eq:orth-expand}
\end{equation}
\begin{equation}
c_{k}(\theta)=\left\langle \Upsilon_{\theta},p_{k}\right\rangle _{\rho}\label{eq:expand-coef}
\end{equation}
where (a.e.) here means (equal) almost everywhere. If $\Upsilon_{\theta}(x)$
is continuous, then it is point-wise equal. In later context we will
denote both cases by simply equal ``=''.

\section{Evolution of Generalized Fourier Series in Gradient Descent}

\subsection{Gradient of Loss in View of Generalized Fourier Series}

The expansion of gradient of loss function provides insight for gradient
descent algorithm:
\begin{align}
\frac{\partial L(\theta)}{\partial\theta} & =\frac{1}{2}\frac{\partial}{\partial\theta}\sum_{k=0}^{\infty}\left\langle f_{\theta}-y,p_{k}\right\rangle _{\rho}^{2}\nonumber \\
 & =\sum_{k=0}^{\infty}\left(c_{k}(\theta)-\left\langle y,p_{k}\right\rangle _{\rho}\right)\frac{\partial c_{k}(\theta)}{\partial\theta}.\label{eq:Loss-d}
\end{align}

For example, if $f_{\theta}$ is 1-layer ANN ($\theta=\{w_{l},r_{l},b_{l}\}$$_{l=1}^{L}$),
with $x\in[-1,1]$
\begin{equation}
f_{\theta}=\sum_{l=1}^{L}w_{l}\sigma(r_{l}x+b_{l}).\label{eq:1-layer-ann}
\end{equation}

Then
\begin{equation}
\frac{\partial f_{\theta}}{\partial\theta}=\begin{cases}
{\displaystyle \sigma(r_{l}x+b_{l})} & \theta=w_{l}\\
w_{l}\sigma'(r_{l}x+b_{l}) & \theta=b_{l}\\
w_{l}\sigma'(r_{l}x+b_{l})x & \theta=r_{l}
\end{cases}\quad l=1,\ldots,L.\label{eq:f-d}
\end{equation}

The last term in Eq. \eqref{eq:Loss-d} is
\begin{equation}
\frac{\partial c_{k}(\theta)}{\partial\theta}=\left\langle \frac{\mathrm{\partial f_{\theta}}}{\partial\theta},p_{k}\right\rangle _{\rho}.\label{eq:ck-derivative}
\end{equation}

If we further assume that $\sigma$ is smooth in the sense of Lemma
\ref{lem:cheb-coef-cn} or Lemma \ref{lem:cheb-coef-analytical},
then $\frac{\partial f_{\theta}}{\partial\theta}$ will have similar
but possibly slightly degraded smoothness. See Lemma \ref{lem:GFS-approximation}
for general $\rho(x)$. The following analysis are based on the Chebyshev
expansion ($\rho(x)=\left(1-x^{2}\right)^{-1/2}$). For $\rho(x)=1$,
it is the Legendre polynomials expansion, it has the same asymptotics
as Chebyshev expansion (Ref. \citep{wang_ontheconvergence_2012}).

Case 1, $\sigma(x)$ is analytical (in the sense of Lemma \ref{lem:cheb-coef-analytical}),
so $f_{\theta}(x)$ and $\frac{\partial f_{\theta}(x)}{\partial\theta}$
are also analytical.

Let's assume that there is no singularity in a neighborhood of real
axis \footnote{If there is singularity on the real axis, then it will be non-smooth
with certain translation parameter $b_{l}$, i.e. probably not a very
interesting activation function.} in the domain of $\sigma(x)$, the nearest singularity is $\pm\mathrm{i}B$
away from the real axis. In this setup, the Bernstein ellipse parameter
$\Theta$ (the $\rho$ in Lemma \ref{lem:cheb-coef-analytical}, not
to be confused with the weighting $\rho(x)$) is $\Theta_{\sigma}=B+\sqrt{1+B^{2}}$
at least.

So the Bernstein ellipse parameter for $\frac{\partial f_{\theta}}{\partial\theta}$
is $\Theta_{\theta}=B/r_{l}+\sqrt{1+\left(B/r_{l}\right)^{2}}\approx\Theta_{\sigma}/r_{l}$
(``$\approx$'' when $r_{l}\ll B$), we get
\[
\left|\frac{\partial c_{k}(\theta)}{\partial\theta}\right|<2M\Theta_{\theta}^{-k}\sim2M\left(\frac{r_{l}}{\Theta_{\sigma}}\right)^{k},\quad k\geq1.
\]

Case 2, $\sigma(x)$ is continuous up to $(\nu-1)$-th derivative
(in the sense of Lemma \ref{lem:cheb-coef-cn}), so $f_{\theta}(x)$
and $\frac{\partial f_{\theta}(x)}{\partial\theta}$ are also smooth
in the same sense, with $\tilde{\nu}=\nu-1$ for $\theta=b_{l},r_{l}$,
$\tilde{\nu}=\nu$ for $\theta=w_{l}$. Due to Lemma \ref{lem:cheb-coef-cn},
we have
\[
\left|\frac{\partial c_{k}(\theta)}{\partial\theta}\right|\leq\frac{2V}{\pi(k-\nu)^{\tilde{\nu}+1}}.
\]

Depending on the curve $\sigma(x)$, the total variation $V$ of $\frac{\mathrm{d}^{\tilde{\nu}}}{\mathrm{d}x^{\tilde{\nu}}}\frac{\partial f_{\theta}}{\partial\theta}$
in $[-1,1]$ can have different asymptotic behavior with respect to
parameter $\theta$. For Relu $\sigma(x)=\max(0,x)$, $\nu=1$, $\mathrm{TV}\left(\sigma^{(\nu)}(r_{l}x)\right)=r_{l}$,
assume the turning point $-b_{l}/r_{l}\in[-1,1]$,
\[
V=\begin{cases}
r_{l} & \theta=w_{l}\\
w_{l}r_{l} & \theta=b_{l}\\
w_{l}(r_{l}+\max(0,2b_{l})) & \theta=r_{l}
\end{cases}.
\]

\subsection{Gradient Descent}

The trajectory of $\theta(t)$ in gradient descent (GD) algorithm
can be rewritten in ODE as a limit of small step size (with or without
$\alpha$, depending on how you discretize the ODE):

\begin{equation}
\frac{\mathrm{d}\theta}{\mathrm{d}t}=-\alpha\left[\frac{\partial L(\theta)}{\partial\theta}\right]^{T},\label{eq:theta-time}
\end{equation}
where $\alpha$ is the learning rate, $\cdot^{T}$ means transpose
of the Jacobian matrix inside, and
\[
\left[\frac{\partial L(\theta)}{\partial\theta}\right]\triangleq\begin{bmatrix}\frac{\partial L(\theta)}{\partial\theta_{1}} & \frac{\partial L(\theta)}{\partial\theta_{2}} & \cdots\end{bmatrix}.
\]

The dynamics of the GFS coefficients $c_{k}(\theta(t))$ can be derived
with help of Eq. \eqref{eq:Loss-d} and Eq. \eqref{eq:theta-time}:
\begin{align}
\frac{\mathrm{d}(c_{k}(\theta(t))-\hat{y}_{k})}{\mathrm{d}t} & =\frac{\partial c_{k}(\theta)}{\partial\theta}\cdot\frac{\mathrm{d}\theta}{\mathrm{d}t}\nonumber \\
 & =-\alpha\sum_{k'=0}^{\infty}\left[\frac{\partial c_{k}(\theta)}{\partial\theta}\right]\left[\frac{\partial c_{k'}(\theta)}{\partial\theta}\right]^{T}(c_{k'}(\theta)-\hat{y}_{k'}),\quad k=0,1,\ldots.\label{eq:coef-dym}
\end{align}

Note that, formally, a system of linear equation
\begin{equation}
\frac{\mathrm{d}\vec{r}}{\mathrm{d}t}=-\alpha H\vec{r},\label{eq:linear-dym}
\end{equation}
with constant symmetric matrix $H$ and spectral decomposition $H=P\Lambda P^{-1}$
has solution:
\begin{equation}
\vec{r}(t)=e^{-\alpha Ht}\vec{C}=Pe^{-\alpha\Lambda t}P^{-1}\vec{C},\label{eq:linear-solution}
\end{equation}
where $\vec{C}$ is an arbitrary constant vector.

It is clear that the eigenvalues of $H$ govern the dynamics of Eq.\eqref{eq:linear-dym}.

With substitution
\begin{equation}
\vec{r}(t)=\left[c_{k}(\theta(t))-\hat{y}_{k}\right]_{k\in\mathbb{N}_{0}},
\end{equation}
\begin{equation}
H=H_{c(\theta)}\triangleq\left[\frac{\partial c_{k}(\theta)}{\partial\theta}\right]_{k\in\mathbb{N}_{0}}\left[\frac{\partial c_{k}(\theta)}{\partial\theta}\right]_{k\in\mathbb{N}_{0}}^{T},\label{eq:H-evolve-matrix}
\end{equation}
where $\left[\cdot\right]_{k\in\mathbb{N}_{0}}$ means vertically
stack the terms indexed by $k$ to form a matrix. e.g.
\[
\left[\frac{\partial c_{k}(\theta)}{\partial\theta}\right]_{k\in\mathbb{N}_{0}}\triangleq\begin{bmatrix}\frac{\partial c_{0}(\theta)}{\partial\theta_{1}} & \frac{\partial c_{0}(\theta)}{\partial\theta_{2}} & \cdots\\
\frac{\partial c_{1}(\theta)}{\partial\theta_{1}} & \frac{\partial c_{1}(\theta)}{\partial\theta_{2}}\\
\vdots &  & \ddots
\end{bmatrix}.
\]

We see the study of eigenvalues could greatly helps the understanding
of Eq.\eqref{eq:coef-dym}, despite $H_{c(\theta)}$ is now time dependent.
Now we turn to the analysis of matrix $H_{c(\theta)}$.

\section{Structure of the matrix $H$}

$H$ is an $\infty\times\infty$ matrix, for simplicity, assume it
is finitely truncated. The truncation is reasonable due to the fast
decaying property of $c_{k}(\theta)$ . Analysis of the infinite dimension
problem directly is under consideration.

Now we need to know some details of $\Upsilon_{\theta}(x)$ for the
investigation of matrix $H$.

\subsection{1-Hidden Layer DNN}

The 1-Hidden Layer DNN with parameter $\theta=\{a_{l},w_{l},b_{l}\}$$_{l=1}^{L}$
and input range $x\in[-1,1]$ is: 

\begin{equation}
\Upsilon_{\theta}(x)\triangleq\sum_{l=1}^{L}a_{l}\sigma(w_{l}x+b_{l}).\label{eq:1-layer-ann-1}
\end{equation}

Due to \eqref{eq:ck-derivative}, matrix $H$ of Eq.\eqref{eq:H-evolve-matrix}
is computed from derivatives of $\Upsilon_{\theta}$ :
\begin{equation}
\frac{\partial\Upsilon_{\theta}(x)}{\partial\theta}=\begin{cases}
{\displaystyle \sigma(w_{l}x+b_{l})} & \theta=a_{l}\\
a_{l}\sigma'(w_{l}x+b_{l}) & \theta=b_{l}\\
a_{l}\sigma'(w_{l}x+b_{l})x & \theta=w_{l}
\end{cases}\quad l=1,\ldots,L.\label{eq:upsilon-derivative}
\end{equation}

The smoothness of $\sigma(x)$ determines the smoothness of $\frac{\partial\Upsilon_{\theta}}{\partial\theta}$
and hence the decaying speed of $\frac{\partial c_{k}(\theta)}{\partial\theta}$
along $k$. Notably there are two common cases:

\subparagraph{Case 1}

$\sigma(x)$ is analytical in a neighborhood of $[-1,1]$ in complex
plane (in the sense of Lemma \ref{lem:cheb-coef-analytical}), so
$\Upsilon_{\theta}(x)$ and $\frac{\partial\Upsilon_{\theta}(x)}{\partial\theta}$
are analytical also. Let's further assume that there is no singularity
in a neighborhood of the real axis \footnote{If there is singularity on the real axis, then it will be non-smooth
with certain translation parameter $b_{l}$. The estimation of $\Theta$
will be more complicated and problem dependent.} for the domain of $\sigma(x)$ in complex plane. i.e. the nearest
singularity is $\pm\mathrm{i}B$ ($B>0$) away from the real axis.
For example, $\sigma(x)=\textrm{sigmoid}(x)=1/(1+\exp(-x))$, has
singularity $x=\mathrm{i}(\pi+2n\pi)\ \forall n\in\mathbb{Z}$, thus
$B=\pi$.

In this setup, the Bernstein ellipse parameter \footnote{The $\Theta$ is the $\rho$ in lemma \ref{lem:cheb-coef-analytical},
not to be confused with the weighting $\rho(x)$.} is $\Theta_{\sigma(x)}\geq B+\sqrt{1+B^{2}}$.

So the Bernstein ellipse parameter for $\frac{\partial\Upsilon_{\theta}}{\partial\theta}$
is $\Theta_{\sigma(w_{l}x)}=B/w_{l}+\sqrt{1+\left(B/w_{l}\right)^{2}}$,
we get
\begin{align}
\left|\frac{\partial c_{k}(\theta)}{\partial\theta}\right| & <2M\Theta_{\sigma(w_{l}x)}^{-k}=2M\left(B/w_{l}+\sqrt{1+\left(B/w_{l}\right)^{2}}\right)^{-k},\quad k\geq1\label{eq:c-decay-exp}\\
 & \approx2M\left(\Theta_{\sigma(x)}/w_{l}\right)^{-k}\quad(\textrm{when}\ w_{l}\ll B)
\end{align}

\subparagraph{Case 2}

$\sigma(x)$ is continuous up to $(\nu-1)$-th derivative (in the
sense of Lemma \ref{lem:cheb-coef-cn}), so $\Upsilon_{\theta}(x)$
also. And $\frac{\partial\Upsilon_{\theta}(x)}{\partial\theta}$ are
continuous up to $(\tilde{\nu}-1)$-th derivative, with $\tilde{\nu}=\nu-1$
for $\theta=b_{l}\textrm{ and }w_{l}$, or $\tilde{\nu}=\nu$ for
$\theta=a_{l}$. By lemma \ref{lem:cheb-coef-cn}, we have upper bound
\begin{equation}
\left|\frac{\partial c_{k}(\theta)}{\partial\theta}\right|\leq\frac{2V}{\pi(k-\tilde{\nu})^{\tilde{\nu}+1}}\quad(\forall k>\tilde{\nu}).\label{eq:c-decay-power}
\end{equation}

\subsection{Graded Matrix}\label{subsec:Graded-Matrix}

Both Eq.\eqref{eq:c-decay-exp} and Eq.\eqref{eq:c-decay-power} show
the decay of $\left|\frac{\partial c_{k}(\theta)}{\partial\theta}\right|$
with respect to $k$. Thus the matrix Eq.\eqref{eq:H-evolve-matrix}
will also have certain decaying property. Specifically, there exist
symmetric matrix $B$ of bounded spectral norm and diagonal matrix
$D$ such that
\begin{equation}
H=DBD\label{eq:grad-decompose}
\end{equation}
where matrix $D$ can be defined as $D=\mathrm{diag}(\left[\Theta_{\sigma(\max(w_{l})x)}^{-k}\right]_{k\in\mathbb{N}_{0}})$
or $D=\mathrm{diag}(\left[(k+1)^{\min(v)-1}\right]_{k\in\mathbb{N}_{0}})$,
depending on the smoothness of $\sigma(x)$. Or, a more straightforward
way is to define $D=\mathrm{diag}(\left[H_{kk}\right]_{k\in\mathbb{N}})$.
And $B=D^{-1}HD^{-1}$ where pseudo-inverse of $D$ is used if necessary.
In the later definition of $D$, it is easy to show that $D_{kk}=O(\Theta_{\sigma(\max(w_{l})x)}^{-k})$
or $D_{kk}=O(k^{-\min(\tilde{\nu})-1})$. The matrix $H$ with structure
Eq.\eqref{eq:grad-decompose} is called graded matrix.

For matrix that has structure Eq.\eqref{eq:grad-decompose}, due to
Min-max theorem of eigenvalues, we have
\begin{equation}
\lambda_{k}(H)\leq D_{kk}^{2}\lambda_{\textrm{max}}(B)\quad\forall k\in\mathbb{N}\label{eq:H-eigval-decay}
\end{equation}
where $\lambda_{k}(H)$ is the $k$-th eigenvalue of $H$ sorting
in descending order. Due to $H_{jj}=D_{kk}^{2}B_{jj}$, Eq. \eqref{eq:H-eigval-decay}
can also written as (assume $B_{jj}\neq0$)
\begin{equation}
\lambda_{k}(H)\leq H_{jj}\frac{\lambda_{\textrm{max}}(B)}{B_{jj}}\quad\forall k\in\mathbb{N}.
\end{equation}

Another way to derive it is by the Schur--Horn theorem. The sorted
array $\left[H_{r(j)r(j)}\right]_{j\in\mathbb{N}_{0}}$ of diagonal
$\left[H_{jj}\right]_{j\in\mathbb{N}_{0}}$ satisfies
\begin{equation}
\sum_{j=1}^{k}H_{r(j)r(j)}\leq\sum_{j=1}^{k}\lambda_{j}\quad k\in\mathbb{N}\label{eq:Hjj-leq-eigval-j}
\end{equation}
and
\begin{equation}
\sum_{j=1}^{\infty}H_{r(j)r(j)}=\sum_{j=1}^{\infty}\lambda_{j}\label{eq:Hjj-eq-eigval-j}
\end{equation}

Eq. \eqref{eq:Hjj-eq-eigval-j} minus Eq. \eqref{eq:Hjj-leq-eigval-j},
with the fact that $\lambda_{j}\geq0$, $H_{jj}\geq0$, we get
\begin{equation}
\lambda_{k}\leq\sum_{j=k}^{\infty}\lambda_{j}\leq\sum_{j=k}^{\infty}H_{r(j)r(j)}.
\end{equation}

For example, if $\left|\frac{\partial c_{k}(\theta)}{\partial\theta_{j}}\right|\leq C_{0}\Theta^{-k}\ \forall k,j$,
\begin{equation}
\left|H_{jk}\right|\leq C_{0}^{2}L\Theta^{-j-k}
\end{equation}
\begin{equation}
\lambda_{k}\leq C_{0}^{2}L\frac{\Theta^{-2k}}{1-\Theta^{-2}}
\end{equation}

\section{Related Problems That Encounter Similar Matrix $H$}

\subsection{Hessian of Loss Function}

The hessian of loss function is
\begin{equation}
\frac{\partial L(\theta)}{\partial\theta}=\int_{\Omega}(\Upsilon_{\theta}(x)-y(x))\frac{\partial\Upsilon_{\theta}(x)}{\partial\theta}\rho(x)\,\mathrm{d}x
\end{equation}
\begin{equation}
\nabla_{\theta}\otimes\nabla_{\theta}L(\theta)=\int_{\Omega}\left(\left[\frac{\partial\Upsilon_{\theta}(x)}{\partial\theta}\right]^{T}\cdot\left[\frac{\partial\Upsilon_{\theta}(x)}{\partial\theta}\right]+(\Upsilon_{\theta}(x)-y(x))\nabla_{\theta}\otimes\nabla_{\theta}\Upsilon_{\theta}(x)\right)\rho(x)\,\mathrm{d}x
\end{equation}

Define
\begin{equation}
H_{L(\theta)}\triangleq\int_{\Omega}\left[\frac{\partial\Upsilon_{\theta}(x)}{\partial\theta}\right]^{T}\cdot\left[\frac{\partial\Upsilon_{\theta}(x)}{\partial\theta}\right]\rho(x)\,\mathrm{d}x.
\end{equation}

When the $\theta$ reaches the global optimal solution $\theta^{*}$
such that $\Upsilon_{\theta^{*}}(x)\equiv y(x)$, the hessian become
\begin{equation}
\nabla_{\theta}\otimes\nabla_{\theta}L(\theta^{*})=H_{L(\theta^{*})}.
\end{equation}

When writing in terms of GFS,
\[
\left[H_{L(\theta)}\right]_{ij}=\left\langle \frac{\partial\Upsilon_{\theta}(x)}{\partial\theta_{i}},\frac{\partial\Upsilon_{\theta}(x)}{\partial\theta_{j}}\right\rangle _{\rho}=\sum_{k=0}^{\infty}\frac{\partial c_{k}(\theta)}{\partial\theta_{i}}\cdot\frac{\partial c_{k}(\theta)}{\partial\theta_{j}},
\]
we have
\begin{equation}
H_{L(\theta)}=\left[\frac{\partial c_{k}(\theta)}{\partial\theta}\right]_{k\in\mathbb{N}_{0}}^{T}\left[\frac{\partial c_{k}(\theta)}{\partial\theta}\right]_{k\in\mathbb{N}_{0}}.\label{eq:loss-optimal-H}
\end{equation}

The matrix $H_{L(\theta)}$ and $H_{c(\theta)}$ (Eq.\eqref{eq:H-evolve-matrix})
are differed only by the multiplication order of the transposed matrix,
therefore they have exactly the same eigenvalues. For simplicity,
define $\lambda_{k}=0$ when $k$ exceed corresponding matrix dimension.
\begin{equation}
\lambda_{k}(H_{L(\theta)})=\lambda_{k}(H_{c(\theta)})\quad(\forall k\in\mathbb{N})\label{eq:eigval-HL-H}
\end{equation}

Thus from Sec. \ref{subsec:Graded-Matrix} we know that the landscape
of loss is extremely flat due to the vast near zero eigenvalues as
shown in Eq.\eqref{eq:H-eigval-decay}.

\subsection{Effective Dimension of Base Function}

The following equation are arised in Extreme learning machine \citep{huang_extreme_2006},
or fixing weights $w_{l}$, $b_{l}$ fitting only $a_{j}$ in Eq.
.

Define
\[
H_{\Upsilon}=\int_{\Omega}\left[\sigma(w_{l}x+b_{l})\right]_{l=1\cdots L}\left[\sigma(w_{l}x+b_{l})\right]_{l=1\cdots L}^{T}\rho(x)\,\mathrm{d}x,
\]
and
\[
\tilde{H}_{\Upsilon}=\sum_{j=1}^{J}\left[\sigma(w_{l}x_{j}+b_{l})\right]_{l=1\cdots L}\left[\sigma(w_{l}x_{j}+b_{l})\right]_{l=1\cdots L}^{T}.
\]

It is clear that $H_{\Upsilon}$ has the same eigenvalue decaying
property as $H_{c(\theta)}$ or $H_{L(\theta)}$. And $\tilde{H}_{\Upsilon}$....
Ideally, the parameters $\left[a_{l}\right]_{l=1\cdots L}$ can be
solved through
\[
\begin{bmatrix}\sigma(w_{1}x_{1}+b_{1}) & \sigma(w_{2}x_{1}+b_{2}) & \cdots & \sigma(w_{L}x_{1}+b_{L})\\
\sigma(w_{1}x_{2}+b_{1}) & \sigma(w_{2}x_{2}+b_{2}) & \cdots & \sigma(w_{L}x_{2}+b_{L})\\
\vdots &  &  & \vdots\\
\sigma(w_{1}x_{J}+b_{1}) & \sigma(w_{2}x_{J}+b_{2}) & \cdots & \sigma(w_{L}x_{J}+b_{L})
\end{bmatrix}\begin{bmatrix}a_{1}\\
a_{2}\\
\vdots\\
a_{L}
\end{bmatrix}=\begin{bmatrix}y(x_{1})\\
y(x_{2})\\
\vdots\\
y(x_{J})
\end{bmatrix}
\]
by its normal equation
\[
\tilde{H}_{\Upsilon}\begin{bmatrix}a_{1}\\
a_{2}\\
\vdots\\
a_{L}
\end{bmatrix}=\sum_{j=1}^{J}\left[\sigma(w_{l}x_{j}+b_{l})\right]_{l=1\cdots L}y(x_{j})
\]

However, due to the analysis above $\tilde{H}_{\Upsilon}$ is numerically
not full rank. 

\section*{Acknowledgment}

It was a great pleasure to work with the authors of \citep{xu_frequency_2020}.
In Oct. 2018, the fruitful gethering discussion, generate core idea
and initial development of this theory, and we see light that might
comming through the mistory of internal workings of ANN.

\section*{Code availability}

The code to help reproduce the main conslutions of this article is
here \href{https://github.com/bewantbe/ANN_GFS}{https://github.com/bewantbe/ANN\_GFS}
mostly structures to help build orthogonal polynomial representations
and simulting a shallow ANN. An early version of this article can
be found here: \url{https://github.com/bewantbe/dynamics_dnn}, note
the file named \emph{dynamics\_1dnn.lyx}.

\section*{Declaration of generative AI and AI-assisted technologies in the
manuscript preparation process}

AI and AI-assisted technologies is not used in the construction of
the Theory at all. In fact, the theory is constructed before the AI-era
(2023), around 2018, . The AI was used to correct gramma and spellings
of this script. Some AI coding was used to generate Figure xxxx. The
author take full resposability to all content of the article.

\bibliographystyle{elsarticle-num}
\bibliography{refs}

\appendix

\section{Lemmas}

See Ref. \citep{gottlieb_numerical_1977,trefethen_approximation_2013}.
Orthogonal polynomial is an active field in 1910s see Bernstein et
al.

\subsection{Chebyshev Approximations}

\subsubsection{Theorems for absolutely continuous functions.}
\begin{lem}[Theorem 7.1 in \citep{trefethen_approximation_2013}]
 Assume $f,\ldots,f^{(\nu-1)}$ be absolutely continuous on $[-1,1]$
and $f^{(\nu)}$ is of bounded variation $V$. Then for $\forall k\geq\nu+1$,
the Chebyshev coefficients of $f$ satisfy
\[
\left|a_{k}\right|\leq\frac{2V}{\pi k(k-1)\cdots(k-\nu)}\leq\frac{2V}{\pi(k-\nu)^{\nu+1}}.
\]
\label{lem:cheb-coef-cn}
\end{lem}

\begin{lem}[Theorem 7.2 in \citep{trefethen_approximation_2013}]
 The same condition as Lemma \ref{lem:cheb-coef-cn}, then for $\forall n>\nu$,
its Chebyshev projections satisfy
\[
\left\Vert f-f_{n}\right\Vert _{\infty}\leq\frac{2V}{\pi\nu(n-\nu)^{\nu}}
\]
and its Chebyshev interpolants satisfy
\[
\left\Vert f-p_{n}\right\Vert _{\infty}\leq\frac{4V}{\pi\nu(n-\nu)^{\nu}}.
\]
\end{lem}

Theorems for analytical functions.
\begin{lem}[Theorem 8.1 in \citep{trefethen_approximation_2013}]
 Assume $f$ analytic in $[-1,1]$ be analytically continuable to
the open Bernstein ellipse $E_{\rho}$ (note \footnote{Bernstein ellipse $E_{\rho}$: foci 1 and -1 with semimajor and semiminor
axis lengths summing to $\rho$ in complex plane.}), where it satisfies $\left|f(x)\right|\leq M$ for some $M$. Then
its Chebyshev coefficients satisfy $\left|a_{0}\right|\leq M$ and
\[
\left|a_{k}\right|\leq2M\rho^{-k},\quad k\geq1.
\]
\label{lem:cheb-coef-analytical}
\end{lem}

\begin{lem}[Theorem 8.2 in \citep{trefethen_approximation_2013}]
 The same condition as Lemma \ref{lem:cheb-coef-analytical}, then
for $\forall n\geq0$ its Chebyshev projections satisfy
\[
\left\Vert f-f_{n}\right\Vert _{\infty}\leq\frac{2M\rho^{-n}}{\rho-1}
\]
and its Chebyshev interpolants satisfy
\[
\left\Vert f-p_{n}\right\Vert _{\infty}\leq\frac{4M\rho^{-n}}{\rho-1}.
\]
 
\end{lem}

\subsection{General Theorems}

General theorems about polynomial approximations.
\begin{lem}[Theorem 16.1 in \citep{trefethen_approximation_2013}]
 Let $f\in C\left([-1,1]\right)$ have degree $n$ Chebyshev projection
$f_{n}$, Chebyshev interpolant $p_{n}$, and best approximant $p_{n}^{*}$,
$n\geq1$. Then
\[
\left\Vert f-f_{n}\right\Vert _{\infty}\leq\left(4+\frac{4}{\pi^{2}}\log(n+1)\right)\left\Vert f-p_{n}^{*}\right\Vert _{\infty},
\]
\[
\left\Vert f-p_{n}\right\Vert _{\infty}\leq\left(2+\frac{2}{\pi}\log(n+1)\right)\left\Vert f-p_{n}^{*}\right\Vert _{\infty}.
\]
\label{lem:bound-cheb-from-best}
\end{lem}

\subsubsection{Two theorems about near-best approximation.}
\begin{lem}[Theorem 12.1 in \citep{trefethen_approximation_2013}]
 Given $f\in C\left([-1,1]\right)$, let $\rho$ ($1\leq\rho\leq\infty$)
be the parameter of the largest Bernstein ellipse $E_{\rho}$ to which
$f$ can be analytically continued, and let $\{p_{n}\}$ be the interpolants
to $f$ in any sequence of grids $\left\{ x_{k}\right\} $ in $[-1,1]$
such that

\[
\lim_{n\rightarrow\infty}M_{n}^{1/n}=\frac{1}{2},\quad M_{n}=\sup_{x\in[-1,1]}\left|l(x)\right|,\quad l(x)=\prod_{k=0}^{n}(x-x_{k}),
\]
then the errors satisfy
\[
\lim_{n\rightarrow\infty}\left\Vert f-p_{n}\right\Vert _{\infty}^{1/n}=\rho^{-1}.
\]
\label{lem:near-best-condition}
\end{lem}

\begin{lem}[Theorem 15.1 in \citep{trefethen_approximation_2013}]
Let $\Lambda$ be the Lebesgue constant for a linear projection $L$
of $C\left([-1,1]\right)$ onto $\mathcal{P}_{n}$. Let $f$ be a
function in $C\left([-1,1]\right)$, $p=Lf$ the corresponding polynomial
approximant to $f$, and $p^{*}$ the best approximation. Then
\[
\left\Vert f-p\right\Vert _{\infty}\leq(\Lambda+1)\left\Vert f-p^{*}\right\Vert _{\infty}
\]
where the Lebesgue constant is defined as
\[
\Lambda\triangleq\sup_{f}\frac{\left\Vert p\right\Vert _{\infty}}{\left\Vert f\right\Vert _{\infty}}=\sup_{x\in[-1,1]}\sum_{j=0}^{n}\left|l_{j}(x)\right|,\quad l_{j}(x)=\frac{\prod_{k\neq j}(x-x_{k})}{\prod_{k\neq j}(x_{j}-x_{k})}.
\]
\label{lem:bound-away-from-best}
\end{lem}

Lemma \ref{lem:near-best-condition} and Lemma \ref{lem:bound-away-from-best}
imply that $\lim_{n\rightarrow\infty}\Lambda_{n}^{1/n}=1$ concludes
near best convergent for $f\in C\left([-1,1]\right)$.
\begin{lem}
\label{lem:GFS-approximation}GFS approximation

For a general weight function $\rho$, Szeg{ő}'s Theorem 12.7.3
\citep{szego_orthogonal_1975} gives the corresponding convergence
result for expansions in orthonormal polynomials. Assume that $\rho$
satisfies the standard Szeg{ő}-class conditions; a standard sufficient
condition is 
\[
\rho(x)>0\quad\text{almost everywhere},\qquad\int_{-1}^{1}\frac{|\log\rho(x)|}{\sqrt{1-x^{2}}}\,\mathrm{d}x<\infty.
\]
Let $\rho_{\star}$ denote the parameter of the largest Bernstein
ellipse $E_{\rho_{\star}}$ in whose interior $g$ is analytic. Then
the GFS coefficients satisfy 
\begin{equation}
\liminf_{k\to\infty}|c_{k}|^{-1/k}=\rho_{\star}.\label{eq:gfs_analytic_rate}
\end{equation}
Consequently, the GFS converges to $g$ inside $E_{\rho_{\star}}$,
uniformly on every compact subset of the ellipse, and diverges outside
it. Thus, for admissible weight functions, the asymptotic geometric
decay of the GFS coefficients is governed by the same Bernstein ellipse
as in the Chebyshev case. The weight function changes the polynomial
basis and the prefactors, whereas the ellipse of convergence is determined
by the analyticity of $g$.
\end{lem}

\end{document}